\documentclass[11pt,a4paper]{article}
\usepackage[whole]{bxcjkjatype}
\usepackage[hyperref]{naaclhlt2018}
\usepackage{times}
\usepackage{latexsym}
\usepackage{multirow}

\usepackage{url}

\aclfinalcopy 
\def\aclpaperid{3014} 

\title{Japanese Predicate Conjugation for Neural Machine Translation}

\author{Michiki Kurosawa, Yukio Matsumura, Hayahide Yamagishi and Mamoru Komachi\\
Tokyo Metropolitan University\\
Hino city, Tokyo, Japan\\
\{\textit{kurosawa-michiki, matsumura-yukio, yamagishi-hayahide}\}@ed.tmu.ac.jp\\
\textit{komachi}@tmu.ac.jp}

\date{}

\begin{document}
\maketitle
\begin{abstract}
  Neural machine translation (NMT) has a drawback in that can generate only high-frequency words owing to the computational costs of the softmax function in the output layer.\\
  In Japanese-English NMT, Japanese predicate conjugation causes an increase in vocabulary size. For example, one verb can have as many as 19 surface varieties. In this research, we focus on predicate conjugation for compressing the vocabulary size in Japanese. The vocabulary list is filled with the various forms of verbs. We propose methods using predicate conjugation information without discarding linguistic information. The proposed methods can generate low-frequency words and deal with unknown words. Two methods were considered to introduce conjugation information: the first considers it as a token (\textbf{conjugation token}) and the second considers it as an embedded vector (\textbf{conjugation feature}).\\
  The results using these methods demonstrate that the vocabulary size can be compressed by approximately 86.1\% (Tanaka corpus) and the NMT models can output the words not in the training data set. Furthermore, BLEU scores improved by 0.91 points in Japanese-to-English translation, and 0.32 points in English-to-Japanese translation with ASPEC.\\

\end{abstract}

  \begin{table*}[t]
   \centering
   \small
   \begin{tabular}{c||cccccc}
    \hline
    語幹&未然形&連用形&終止形&連体形&仮定形&命令形\\
    Stem&Irrealis&Continuative&Terminal&Attributive&Hypothetical&Imperative\\
    \hline
    \hline
    走る&走ら (\textit{hashi-ra})&走り&走る&走る&走れ&走れ\\
    (\textit{hashi-ru}; run)&走ろ (\textit{hashi-ro})&(\textit{hashi-ri})&(\textit{hashi-ru})&(\textit{hashi-ru})&(\textit{hashi-re})&(\textit{hashi-re})\\
    \hline
    歩く&歩か (\textit{aru-ka})&歩き&歩く&歩く&歩け&歩け\\
    (\textit{aru-ku}; walk)&歩こ(\textit{aru-ko})&(\textit{aru-ki})&(\textit{aru-ku})&(\textit{aru-ku})&(\textit{aru-ke})&(\textit{aru-ke})\\
    \hline
    する&せ (\textit{se})&し&する&する&すれ&しろ (\textit{shi-ro})\\
    (\textit{su-ru}; do)&し (\textit{shi})&(\textit{shi})&(\textit{su-ru})&(\textit{su-ru})&(\textit{su-re})&せよ (\textit{se-yo})\\
    \hline
   \end{tabular}
   \caption{Leverage table of verb.}
   \label{table:run}
  \end{table*}

\section{Introduction}

  Neural machine translation (NMT) is gaining significant attention in machine translation research because it produces high-quality translation \cite{bahdanau2014neural, luong-pham-manning:2015:EMNLP}.
  However, because NMT requires massive computational time to select output words, it is necessary to reduce the vocabulary in practice by using only high-frequency words in the training corpus. Therefore, NMT treated not only \textbf{unknown words}, which do not exist in the training corpus, but also \textbf{OOV}, which can not consider words to NMT's computational ability, as unknown word token\footnote{In this paper, we denote a word not appearing in the training corpus as ``unknown word,'' and a word treated as an unknown low-frequency word as ``OOV.''}.\par
  Two approaches were proposed to address this problem: backoff dictionary \cite{luong-EtAl:2015:ACL-IJCNLP} and byte pair encoding, or BPE \cite{sennrich-haddow-birch:2016:P16-12}. However, because the backoff dictionary is a post-processing method to replace OOV, it is not a fundamental solution. BPE can eliminate unknown words by dividing a word into partial strings; however, there is a possibility of loss of linguistic information such as loss of the meaning of words.\par
In Japanese grammar, the surfaces of verb, adjective, and auxiliary verb change into different forms by the neighboring words.
This phenomenon is called ``conjugation,'' and 18 conjugation patterns can be formed at maximum for each word.
We consider the conjugation forms as the vocabulary of NMT using Japanese language because the Japanese morphological analyzer divides a sentence into words based on conjugation forms.
The vocabulary set in the NMT model must have all conjugation forms for generating fluent sentences.\par
  In this research, we propose two methods using predicate conjugation information without discarding linguistic information. These methods can not only reduce OOV words, but also deal with unknown words. In addition, we consider a method to introduce part-of-speech (POS) information other than predicate. We found this method is related to source head information.\\
  The main contributions of this paper are as follows:
  \begin{itemize}
  \item The proposed NMT reduced the vocabulary size and improved BLEU scores particularly in small- and medium-sized corpora.
  \item We found that conjugation features are best exploited as tokens rather than embeddings and suggested the connection between the position of the token and linguistic properties.
  \end{itemize}

\section{Related work}

\paragraph{Backoff dictionary.}
  \newcite{luong-EtAl:2015:ACL-IJCNLP} proposed a method of rewriting an unknown word token in the output into an appropriate word using a dictionary. This method determines a corresponding word using alignment information between an output sentence and an input sentence and rewrites the unknown word token in the output using the dictionary. Therefore, it does not allow NMT to consider the meaning of OOVs. However, this method can be used together with the proposed method, which results in the further reduction of unknown words.

\paragraph{Byte pair encoding.}
  \newcite{sennrich-haddow-birch:2016:P16-12} proposed a method to construct vocabulary by splitting all the words into characters and re-combining them based on their frequencies to make sub-word unit. Because all words can be split into known words based on characters, this method has an advantage in that OOV words disappear. However, because coupling of subwords depends on frequency, grammatical and semantic information is not taken into consideration. Incidentally, Japanese has many characters especially kanji; therefore, there might exist unknown characters that do not exist in the training corpus even after applying BPE.

\paragraph{Input feature.}
  \newcite{sennrich-haddow:2016:WMT} proposed a method to add POS information and dependency structure as embeddings with the aim of explicitly learning syntax information in NMT. However, it can only be applied to the input side.

\section{Japanese predicate conjugation}

 Japanese predicates consist of stems and conjugation suffixes. In the vocabulary set obtained by conventional word segmentation, they are treated as different words. Therefore, the vocabulary set is occupied with predicates which have similar meaning but different conjugation.\par
 As an example, a three-type conjugation table is shown in Table \ref{table:run}. In this way, conjugation represents many expressions with only a subtle difference in meaning. Due to the Japanese writing system, most of the predicates do not share conjugation suffixes even though they share the same conjugation patterns. Comparing ``走る (run)'' and ``歩く (walk)'', if one wants to share the conjugation suffixes using BPE, it is necessary to represent these words using Latin alphabets instead of phonetic characters, or kana. In addition, a special verb ``する (do)'' cannot share the conjugation suffixes with these words even using BPE. Therefore, we cannot divide the predicates into the stems and shared conjugation suffixes using BPE.\par
 In the proposed method, we handle them collectively. Since types of conjugation are limited, we can deal with every types. All conjugation forms can be consolidated into one lemma, and OOV can be reduced\footnote{Derivational grammar \cite{Ogawa} to unify multiple conjugation forms, but it cannot distinguish between plain and attributive forms and imperfective and continuative forms if they have the same surface.}. Furthermore, by treating a lemma and conjugation forms as independent words, it is possible to represent the predicates which we were observed a few times on the training corpus by combining lemmas and conjugation forms found in the training corpus.\par
 In this research, MeCab\footnote{\url{https://github.com/taku910/mecab}} is used as a Japanese morphological analyzer, and the morpheme information adopts the standard of IPADic. Specifically, ``\textit{surface form}'', ``\textit{POS (coarse-grained)}'', ``\textit{POS (fine-grained)}'', ``\textit{conjugation type}'', ``\textit{conjugation form}'', and ``\textit{lemma}'' are used.
 Hereafter, predicates represent verbs, adjectives, and auxiliary verbs.\par

\section{Introducing Japanese predicate conjugation for NMT}

 We propose two methods to introduce conjugation information: in the first method, it is treated as a token (\textbf{conjugation token}) and in the second, it is treated as concatenation of embeddings (\textbf{conjugation feature}). Moreover we considere to introduce POS information into all words (\textbf{POS token}).

\subsection{Conjugation token}
\label{subsection:conjugation token}
 In this method, lemmas and conjugation forms are treated as tokens. A conjugation form is introduced as a special token with which its POS can be distinguished from other tokens.\par
 In this method, the special token also occupies a part of the vocabulary. However, as there are only 55 tokens\footnote{Verb: 19, Adjective: 14, Auxiliary verb: 22} at maximum in the IPADic standard, the influence is negligible compared to the vocabulary size that can be reduced. Moreover, because the stem and its conjugation suffix are explicitly retrieved, the output can be restored at any time.\\
 For example, these are converted as follows.
 
 \begin{center}
 \small
 \begin{tabular}{cccc}
  走る&→&走る&\verb#<動詞・基本形>#\\
  &&(run)&(verb--plain)\\
  走れ&→&走る&\verb#<動詞・命令形>#\\
  &&(run)&(verb--imperative)\\
  だ&→&だ&\verb#<助動詞・体言接続>#\\
  &&(COPULA)&(aux.verb--attributive)\\
 \end{tabular}
 \end{center}
 \normalsize

\subsection{Conjugation feature}
\label{subsection:conjugation feature}
 In this method, we use a conjugation form as a feature of input side. Specifically, ``\textit{POS (coarse-grained)}'', ``\textit{POS (fine-grained)}'', and ``\textit{conjugation forms}'' are used in addition to the lemma. Moreover, this information is added to words other than predicates. These features are first represented as one-hot vectors, and the learned embedding vectors are concatenated and used.\par
 This method has an advantage in that it does not waste vocabulary size; however, because it is not trivial to restore a word from embeddings, it can be adopted to the source side only.

\subsection{POS token}
 As a natural extension to Conjugation token, we introduce POS information into all words in addition to conjugation information. We use POS information and conjugation information in the same manner to Conjugation token. We propose three methods to incorporate POS information as special tokens.\par

 \paragraph{Suffix token.}
 \label{subsection:suffix}
 This method introduces POS and conjugation information behind each word as a token.
 
 \paragraph{Prefix token.}
 \label{subsection:prefix}
 This method introduces POS and conjugation information in front of each word as a token.
 
 \paragraph{Circumfix token.}
 \label{subsection:circumfix}
 This method introduces POS information in front of each word and conjugation information behind each word as a token.\\\par

Example sentences are shown below:

 \begin{description}
 \item[Baseline]\mbox{}\\ \small 私 は 走る 。(I run .) \normalsize
 \item[Suffix token]\mbox{}\\ \small 私 \verb|<noun>| は \verb|<particle>| 走る \verb|<verb-plain>| \verb|<verb>| 。 \verb|<symbol>| \normalsize
 \item[Prefix token]\mbox{}\\ \small \verb|<noun>| 私 \verb|<particle>| は \verb|<verb>| \verb|<verb-plain>| 走る \verb|<symbol>| 。 \normalsize
 \item[Circumfix token]\mbox{}\\ \small \verb|<noun>| 私 \verb|<particle>| は \verb|<verb>| 走る \verb|<verb-plain>| \verb|<symbol>| 。\normalsize
 \end{description}

    \begin{table}[t]
   \centering
   \small
   \begin{tabular}{l|rrrc}
    \hline
    Corpus&\multicolumn{1}{c}{train}&\multicolumn{1}{c}{dev}&\multicolumn{1}{c}{test}&Max length\\
    \hline
    \hline
    NTCIR&1,638,742&2,741&2,300&60\\
    ASPEC&827,503&1,790&1,812&40\\
    Tanaka&50,000&500&500&16\\
    \hline
   \end{tabular}
   \caption{Details of each corpus.}
   \label{table:corpus}
  \end{table}

  \begin{table*}[tp]
   \centering
   \small
   \begin{tabular}{c|l|ccc|ccc}
   \hline
   \multicolumn{2}{c|}{\multirow{2}{*}{Method}}&\multicolumn{3}{c|}{Japanese - English}&\multicolumn{3}{c}{English - Japanese}\\
   \multicolumn{2}{c|}{}&NTCIR&ASPEC&Tanaka&NTCIR&ASPEC&Tanaka\\
   \hline
   \hline
   \multirow{4}{*}{Baseline}&w/o BPE&33.87&20.98&30.23&36.41&29.57&30.25\\
   &BPE only Japanese&\textbf{34.17}&21.10&30.43&35.96&28.96&28.66\\
   &BPE both sides&-&21.43&30.45&-&30.93&29.27\\
   &BPE only English&-&20.55&30.13&-&30.59&29.15\\
   \hline
   \multirow{2}{*}{\shortstack{Only predicate\\conjugation information}}&(\ref{subsection:conjugation token}) Conjugation token&33.96&21.47&32.47&\textbf{36.48}&\textbf{29.89}&30.46\\
   &(\ref{subsection:conjugation feature}) Conjugation feature&33.84&21.33&30.35&N/A&N/A&N/A\\
   \hline
   \multirow{3}{*}{\shortstack{Using predicate\\conjugation information\\and all POS information}}&(\ref{subsection:suffix}) Suffix token&-&21.49&31.82&-&29.77&\textbf{31.47}\\
   &(\ref{subsection:prefix}) Prefix token&-&21.61&32.16&-&29.02&30.36\\
   &(\ref{subsection:circumfix}) Circumfix token&-&\textbf{21.89}&\textbf{32.96}&-&28.89&31.07\\
   \hline
   \end{tabular}
   \caption{BLEU scores of each experiment (average of four runs). The best score in each corpus is made bold (expect for BPE ``both'' and ``only English'').}
   \label{table:result}
  \end{table*}

\section{Experiment}

 We experimented two baseline methods (with and without BPE) and two proposed methods. Each experiment was conducted four times with different initializations. We report the average performance over all experiments.\par
 We used three data sets: NTCIR PatentMT Parallel Corpus - 10 \cite{NTCIR}, Asian Scientific Paper Excerpt Corpus \cite{NAKAZAWA16.621}, and Tanaka Corpus (Excerpt, Preprocessed)\footnote{\url{http://github.com/odashi/small_parallel_enja}}. The details of each corpus are shown in Table \ref{table:corpus}. Only in Tanaka, English sentences were already lowercased; hence, truecase was not used. As for ASPEC, we used only the first one million sentences sorted by sentence alignment confidence. Japanese sentences were tokenized by the morphological analyzer MeCab (IPADic), and English sentences were preprocessed by Moses\footnote{\url{http://www.statmt.org/moses/}} (tokenizer, truecaser). As for the training corpus, we deleted sentences that exceeded the maximum number of tokens each sentence shown in Table \ref{table:corpus}.\par
 We used our implementation\footnote{\url{http://github.com/yukio326/nmt-chainer}} based on \newcite{luong-pham-manning:2015:EMNLP}
  as the baseline. Hyper-parameters are as follows. If the setting differs in the corpus, it is written in the order of NTCIR / ASPEC / Tanaka.\par
   Optimization: AdaGrad, Learning rate: 0.01,\par
   Embed size: 512, Hidden size: 1,024,\par
   Batch size: 128, Maximum epoch: 15 / 15 / 30,\par
   Vocab size: 30,000 / 30,000 / 5,000, \par Output limit: 100 / 100 / 40\\
The setting of each experiment except the baseline is shown below. We used the same setting as the baseline unless otherwise specified.

\paragraph*{Byte pair encoding.}
 We conducted an experiment using BPE as the comparative method. BPE was applied to the Japanese side only for making a fair comparison with the proposed method.\par
 The number of merge operations in both NTCIR and ASPEC was set to 16,000 and in Tanaka, the number was set to 2,000. As a result, OOV did not exist in all corpora because the size of Japanese vocabulary is smaller than that of BPE.

\paragraph*{Conjugation token.}
Because the output of English--Japanese translation includes special tokens, we evaluate it by restoring the results with rules using IPADic. The restoration accuracy is 100\%. If the output has only a lemma, it is converted into the plain form, and if it has a conjugation token only, the token is deleted from the output.

\paragraph*{Conjugation feature.}
 Because this method can solely be adopted to the source side, only Japanese-to-English translation was performed. To restrict the embed size to 512, the size of each feature was set to POS (coarse-grained): 4, POS (fine-grained): 8, conjugation form: 8, lemma: 492.
 
\paragraph*{POS token.}
 We increased the output limit by 2.5 times in English-to-Japanese translation because of additional POS tokens attached to all words.\par
 We used the same restoration rules as for Conjugation token to treat special tokens.\par
 We evaluated POS features in only ASPEC and Tanaka owing to time constraints.

\section{Discussion}

\subsection{Translation quality}
 The results of BLEU score \cite{papineni-EtAl:2002:ACL} are shown in Table \ref{table:result}. Compared to the baseline without BPE, Conjugation token improved in BLEU score on all corpora and in both translation directions. In addition, Conjugation token outperformed the baselines with BPE with an exception on NTCIR in Japanese-to-English direction. 
 When the POS token was introduced, BLEU scores improved by 1.82 points on average from the baseline in Japanese-to-English translation. (ASPEC : 0.91, Tanaka : 2.73)\par
 Furthermore, we compared proposed methods with the baseline that adopted BPE to the Japanese side only\footnote{Owing to time limitations, we performed comparison with ASPEC and Tanaka corpora only, and experimented only once on each corpus.}. Table \ref{table:result} shows the results of baseline with BPE to both English and Japanese sides. According to the results, Japanese-only BPE was inferior to the baseline without BPE.

  \begin{table}[tp]
   \centering
   \small
   \begin{tabular}{c|ccc}
    \hline
    Corpus&Baseline&\shortstack{Conjugation\\token}&\shortstack{Conjugation\\feature}\\
    \hline
    \hline
    NTCIR&26.48\%&27.43\%&27.46\%\\
    ASPEC&18.56\%&18.96\%&18.96\%\\
    Tanaka&46.46\%&53.95\%&54.41\%\\
    \hline
   \end{tabular}
   \caption{Vocabulary coverage.}
   \label{table:word}
  \end{table}
 
\subsection{Vocabulary coverage}
 The proposed method is effective in reducing the vocabulary size. The coverage of each training corpus is shown in Table \ref{table:word}. As for Conjugation feature, we evaluate only the number of lemmas.\par
 It can be seen that OOV is reduced in all corpora. In particular, a significant improvement was found in the small Tanaka corpus. It can partly account for the improvement in BLEU scores in the proposed methods.
 
\subsection{Effect of conjugation information}
 Experimental results showed that Conjugation token improved the BLEU score.
However, Conjugation feature exhibited little or no improvement over the baselines with and without BPE. It was shown that conjugation information consists of useful features, but we should exploit the information as Conjugation token.\par
 In the Conjugation token method, we found that the scores are influenced by the corpus size. In particular, the largest improvement was seen in a small Tanaka corpus. Conversely, Conjugation token had a small effect in a large NTCIR corpus, where both proposed methods were inferior compared to the baseline using BPE in Japanese-to-English translation. This is because the size of the corpus was sufficient to learn frequent words to produce fluent translations. Also, our method is superior to BPE in small corpus because it can compress the vocabulary without relying on frequency.

\subsection{Output example}
 Tables \ref{table:outputa} and \ref{table:outputb} show the output examples in Japanese-to-English translation results.\par
 Table \ref{table:outputa} depicts the handling of OOV. The baseline without BPE treated ``古来'' (ever lived) in this source sentence as OOV, so it could not translate the word. However, BPE and Conjugation token could translate it because it was included in each vocabulary.\par
 Table \ref{table:outputb} shows the handling of an unknown word. In the baseline without BPE, ``下ろす'' (take down) in the source sentence was represented as an unknown word token because it did not appear on the training corpus, and therefore, it failed to generate ``take down'' correctly. However, the conjugation token could successfully translate it because the lemma (``下ろす'') which appears on the training corpus as the conditional form (``下ろせ''), continuative form (``下ろし''), and plain form (``下ろす'') could be used to generate the plain form \mbox{(``下ろす'')}.

  \begin{table}[tp]
   \centering
   \small
   \begin{tabular}{c|c}
    src&彼 は {\bf 古来}\footnotemark[10] まれ な 大 政治 家 で ある 。\\
    ref&he is as great a statesman as ever lived .\\
    w/o BPE&he is as great a statesman as any .\\
    BPE&he is as great a statesman as ever lived .\\
    C\_token\footnotemark[9]&he is as great a statesman as ever lived .\\
   \end{tabular}
   \caption{Output example 1.}
   \label{table:outputa}
  \end{table}
  
  \begin{table}[tp]
  \centering
  \small
  \begin{tabular}{c|c}
    src&これ を {\bf 下ろす}\footnotemark[10] の てつだっ て ください 。\\
    ref&please give me help in taking this down .\\
    w/o BPE&please take this for me .\\
    BPE&please take this to me .\\
    C\_token\footnotemark[9]&please take this down .\\
   \end{tabular}
   \caption{Output example 2.}
   \label{table:outputb}
  \end{table}

\footnotetext[9]{Abbreviation for Conjugation token.}
\footnotetext[10]{OOV or unknown word in the baseline.}

 \subsection{Effect of POS information}
 Experimental results showed that the Circumfix token (\ref{subsection:circumfix}) achieved the best score in Japanese-to-English translation, whereas the Conjugation token (\ref{subsection:conjugation token}) or suffix token (\ref{subsection:suffix}) was the best in English-to-Japanese translation.\par
 We suppose that the reason for this tendency derives from the head-directionality of the target language. Because the target language in English-to-Japanese translation is Japanese, which is a head-final language, the POS token as the suffix seems to improve the translation accuracy more than the others. \par
However, experimental results in Japanese-to-English translation contradict this hypothesis. We assume that it is because of the right-hand head rule \cite{righthand} in English. According to this rule, basic linguistic information should be introduced before a word whereas inflection information should be placed after the word. This accounts for the different tendency in the performance of the POS token. 

\section{Conclusion}
 In this paper, we proposed two methods using predicate conjugation information for compressing Japanese vocabulary size. The experimental results confirmed improvements in both vocabulary coverage and translation performance by using Japanese predicate conjugation information. It is important for the NMT systems to retain the grammatical property of the target language when injecting linguistic information as a special token. Moreover, it was confirmed that the proposed method is effective not only for OOV but also for unknown words.

\bibliography{reference}
\bibliographystyle{acl_natbib}

\end{document}